\documentclass[lettersize,journal]{IEEEtran}
\usepackage{amsmath,amsfonts}
\usepackage{algorithmic}
\usepackage{algorithm}
\usepackage{array}
\usepackage{textcomp}
\usepackage{stfloats}
\usepackage{url}
\usepackage{verbatim}
\usepackage{graphicx}
\usepackage{hyperref}
\usepackage{cleveref}
\usepackage[nottoc]{tocbibind}
\usepackage{booktabs}
\usepackage{multirow}
\usepackage{tabularx}
\usepackage{caption}
\usepackage{subcaption}
\usepackage{float}
\usepackage{ulem} 
\usepackage{bm}
\usepackage{soul}
\usepackage{xcolor}
\sethlcolor{cyan} 

\crefname{figure}{Figure}{Figure}
\Crefname{figure}{Figure}{Figure}
\crefname{table}{Table}{Table}
\Crefname{table}{Table}{Table}

\newcommand{\argmin}[1]{\underset{#1}{\arg\min}}
\begin{document}

\title{HyDRA: A Hybrid Dual-Mode Network \\ for Closed- and Open-Set RFFI with Optimized VMD}

\author{
    \IEEEauthorblockN{Hanwen Liu\textsuperscript{\#}}
    ,
    \IEEEauthorblockN{Yuhe Huang\textsuperscript{\#}}
    ,
    \IEEEauthorblockN{Yifeng Gong\textsuperscript{\#}}
    ,
    \IEEEauthorblockN{Yanjie Zhai\textsuperscript{\#}}
    ,
    \IEEEauthorblockN{Jiaxuan Lu\textsuperscript{*}}
\thanks{\textsuperscript{\#} Contributed equally to this work.} 
\thanks{Hanwen Liu and Yifeng Gong are with the School of Electronics and Communication Engineering, Sun Yat-sen University, Shenzhen, 518107, China, e-mail: (liuhw56, gongyf9)@mail2.sysu.edu.cn.}
\thanks{Yuhe Huang is with the School of Electronic Information and Electrical Engineering, Shanghai Jiao Tong University, Shanghai, 200240, China, e-mail: yuhehuang@sjtu.edu.cn.}
\thanks{Yanjie Zhai is with the School of Electronic Information and Electrical Engineering, Dalian Maritime University, Dalian, 116026, China, e-mail: yzhai4@outlook.com.}
\thanks{Jiaxuan Lu is with Shanghai Artificial Intelligence Laboratory, Shanghai, 200232, China, e-mail: lujiaxuan@pjlab.org.cn.}
\thanks{* Corresponding Author.}

}

\markboth{Journal of \LaTeX\ Class Files,~Vol.~14, No.~8, August~2021}%
{Shell \MakeLowercase{\textit{et al.}}: A Sample Article Using IEEEtran.cls for IEEE Journals}


\maketitle

\begin{abstract}
Device recognition is vital for security in wireless communication systems, particularly for applications like access control. Radio Frequency Fingerprint Identification (RFFI) offers a non-cryptographic solution by exploiting hardware-induced signal distortions. This paper proposes HyDRA, a Hybrid Dual-mode RF Architecture that integrates an optimized Variational Mode Decomposition (VMD) with a novel architecture based on the fusion of Convolutional Neural Networks (CNNs), Transformers, and Mamba components, designed to support both closed-set and open-set classification tasks. The optimized VMD enhances preprocessing efficiency and classification accuracy by fixing center frequencies and using closed-form solutions. HyDRA employs the Transformer Dynamic Sequence Encoder (TDSE) for global dependency modeling and the Mamba Linear Flow Encoder (MLFE) for linear-complexity processing, adapting to varying conditions. Evaluation on public datasets demonstrates state-of-the-art (SOTA) accuracy in closed-set scenarios and robust performance in our proposed open-set classification method, effectively identifying unauthorized devices. Deployed on NVIDIA Jetson Xavier NX, HyDRA achieves millisecond-level inference speed with low power consumption, providing a practical solution for real-time wireless authentication in real-world environments. {The source code is published on https://github.com/Crazy-Bull/HyDRA.}
\end{abstract}


\section{Introduction}
Radio Frequency Fingerprint Identification (RFFI) identifies wireless devices by analyzing unique hardware-induced signal distortions caused by manufacturing variations in analog components\cite{RFFI_1st}. These distortions manifest as subtle spectral and temporal signatures, enabling device authentication independent of cryptographic protocols. As wireless networks grow in complexity, RFFI has become indispensable for securing IoT deployments\cite{merchant2018deep}\cite{loT_1st}\cite{loT_2ed}, drone communications\cite{zhao2023drone}\cite{Drone_2ed}, and mission-critical systems\cite{vaswani2017attention}\cite{MoreMission_Liu2009}. 

RFFI methodologies are categorized into expert feature-driven approaches and deep learning (DL)-based. Expert methods rely on manually engineered features from time-frequency analysis, such as Hilbert-Huang transforms\cite{Hilbert–Huang} for transient distortion characterization, wavelet decompositions\cite{wireless-transmitters} for steady-state artifact quantification, local mean decomposition (LMD)\cite{LMD} for AM/FM signal decomposition and time-varying instantaneous frequency extraction. While interpretable, these approaches suffer from noise sensitivity under low SNR conditions and limited capability to model nonlinear hardware impairment interactions. Their reliance on heuristic feature design further restricts scalability in congested spectral environments\cite{Rehman2012TheAO}.

The advent of Deep neural network (DNN) shifted focus toward automated feature extraction\cite{DNN}\cite{CRD_comparasion}, including Recurrent Neural Network (RNN)\cite{RNN_Wang2018} and Convolutional Neural Network (CNN)\cite{Zigbee_CNN}\cite{CNN_Constellation}. Existing DL-based methods utilize two parallel paradigms: transient and steady-state analysis. Transient-based approaches leverage brief signal transitions to capture hardware fingerprints\cite{transient}. However, they require expensive high-speed sampling equipment\cite{TransientEC} and degrade severely under mobility-induced Doppler shifts\cite{ModularBluetooth}. These limitations are particularly acute in WiFi systems, where rare device reboots result in scarce transient data. In contrast, steady-state methods analyze persistent signal segments like protocol preambles\cite{Vector_Steady}\cite{Steady_State_ZHAO2017164}, which provide deterministic structures unaffected by payload randomness. This enables robust multi-scale periodicity analysis and efficient channel equalization, making steady-state approaches preferable for real-world deployment.

To extract feature of steady-state signal, CNN-based models\cite{CNN_1st}\cite{Channel_aware} effectively capture hardware-specific features. These models demonstrate strong performance in scenarios with limited devices and short signal segments. However, the inherent limitation of CNNs in modeling global dependencies due to their local receptive fields leads to significant performance degradation when applied to large-scale datasets. To optimize computational efficiency, ADCC\cite{CNN_2ed} employ dilated convolutions and signal truncation preprocessing, reducing computational complexity for edge deployment. However, these optimizations sacrifice time-space resolution, demonstrating that dilated convolutions expand the receptive field but reduce sensitivity to signal variations, while signal truncation may discard device-specific features in later signal segments\cite{Few_shot}. 

Optimized preprocessing methods also contribute on hardware-specific feature and computational efficiency. Zhang et al. \cite{Hilbert–Huang} utilized energy entropy and Hilbert spectra derived from Empirical Mode Decomposition (EMD)\cite{EMD} for classification, while Flamant et al. \cite{Flamant2018} analyzed time-frequency characteristics of bivariate signals. Dragomiretskiy et al. \cite{dragomiretskiy2013variational} demonstrate that Variational Mode Decomposition (VMD) is adaptive and noise-resistant and can effectively reduce the complexity of time-frequency feature extraction. Despite these advancements, limitations persist: EMD and bivariate characteristics suffer from endpoint effects and frequency aliasing, while the inherent reconstruction artifacts in conventional VMD frameworks progressively compromise spectral fingerprint discernibility through error accumulation in large-scale datasets {by applying Alternating Direction Method of Multipliers (ADMM). It introduces reconstruction error from augmented Lagrangian terms, which filters out subtle hardware-induced signal distortions that are the core of RFFI}; additionally, the computational complexity of VMD models imposes significant resource demands, with these limitations escalating exponentially as the number of Intrinsic Mode Functions (IMFs) channels increases.

For the testing method, current research on RFFI predominantly focuses on closed-set testing, where models are trained and evaluated solely on predefined known device categories\cite{Openset_Xie2021}. However, this approach starkly conflicts with real-world security requirements. In practical wireless communication environments, systems must simultaneously achieve precise identification of authorized devices and effective interception of unauthorized ones. Traditional closed-set models, by forcing all test samples into predefined categories, risk misclassifying unknown devices as legitimate, posing critical security vulnerabilities.

To address these challenges, we propose the HyDRA framework, which integrates lossless Variational Mode Decomposition (VMD) and a reconfigurable dual-encoder architecture to reduce computational time, avoid convergence loss, balance accuracy and efficiency, and enable open-set testing. The architecture dynamically switches between the Transformer Dynamic Sequence Encoder (TDSE) for high-precision tasks and the Mamba Linear Flow Encoder (MLFE) for high-efficiency demands, with {the TDSE uses multi-head self-attention to model global dependencies across all signal time steps. This allows the model to capture cross-segment correlations between hardware distortions, a capability critical for large-scale device identification, where subtle fingerprint differences often span long signal sequences; The MLFE leveraging a selective state-space model (SSM) to use input-dependent parameters that selectively focus on distortion-rich signal segments, enabling long-range dependency capture at linear computational complexity. This is theoretically superior to traditional Transformers, as it avoids redundant pairwise attention calculations—making it feasible for resource-constrained edge devices while retaining long-range feature learning ability.} This dual-mode design addresses the diverse requirements of real-world wireless communication systems, where high-precision identification is critical in secure, high-stakes environments, while rapid, energy-efficient processing is essential for resource-constrained edge devices. Channel-aware feature fusion is enhanced through split-path residual blocks that decouple magnitude and phase processing in the RF domain, using adaptive concatenation weights adjusted by a Convolutional Feature Refinement Extractor (CFRE) to extract multi-scale temporal features. We further redesign VMD by fixing central frequencies and deriving closed-form expressions for Intrinsic Mode Function (IMF) components, eliminating the iterative ADMM to minimize computational time and reconstruction losses. For open-set testing, we introduce a threshold discrimination-based classification framework that utilizes the model’s softmax probability distribution to assess device identity reliability, flagging devices as unauthorized if the highest probability falls below a preset threshold, thereby enhancing robustness in real-world scenarios. Our contributions are:
\begin{itemize}
\item We propose the HyDRA dual-mode framework, integrating optimized lossless VMD with a CNN-Transformer-Mamba network, achieving SOTA accuracy in closed-set testing on the WiSig dataset\cite{Wisig2022} and pioneering open-set testing.

\item We introduce a dynamic dual-encoder architecture: TDSE enabling global modeling under high SNR conditions, MLFE using selective state space models for linear complexity, achieving millisecond-level speed with low power consumption on NVIDIA Jetson Xavier NX.

\item A data processing method is developed by fixing VMD central frequencies and using closed-form solutions to eliminate ADMM reconstruction error. This method achieves higher accuracy on the WiSig dataset\cite{Wisig2022} compared to raw IQ data and VMD (ADMM), while significantly reducing decomposition time compared to VMD (ADMM).

\item A threshold discrimination-based open-set classification mechanism is proposed, achieving a peak accuracy of 94.67\% in detecting illegal devices during open-set testing, enhancing RFFI robustness in real-world scenarios.
\end{itemize}

\section{Related Works}
\subsection{Spectral Feature-Based RFFI}
Early works established foundational techniques for hardware fingerprint extraction through time-frequency analysis. Zhang et al.~\cite{Hilbert–Huang} pioneered Hilbert-Huang transforms to characterize transient signal distortions, while Dragomiretskiy et al.~\cite{dragomiretskiy2013variational} introduced VMD to decompose signals into narrowband IMFs. Subsequent studies explored dual-tree wavelet transforms~\cite{wireless-transmitters} for multi-resolution analysis and I/Q imbalance metrics~\cite{Zhuo2019} for steady-state feature extraction.  These methods achieved promising results in controlled scenarios but faced limitations in dynamic environments due to fixed decomposition parameters, which restricted their adaptability to channel variations and hardware diversity. Rehman et al. \cite{Rehman2012TheAO} proposed utilizing the spectrogram of a short-time Fourier transform to obtain the energy envelope of instantaneous transient signals and extract unique features from the envelope for validation through Bluetooth signal recognition. Recent advancements by Liang et al.~\cite{WANG2022110798} developed adaptive VMD through reinforcement learning, dynamically adjusting decomposition parameters based on channel state information. Han et al.~\cite{Han2025} integrated VMD with Transformer encoders, achieving 67.83\% recognition accuracy under mobility conditions—a critical step toward adaptive decomposition frameworks.

\subsection{Deep Learning-Based RFFI}
Deep learning paradigms revolutionized RFFI through automated feature learning. Transient analysis leverages temporal dynamics of signal activation states. For example, CNNs~\cite{Zigbee_CNN} achieved 93\% accuracy in classifying ZigBee power-on signatures by capturing abrupt waveform transitions, while recurrent networks~\cite{Enhanced_Networks} modeled sequential dependencies in transient phases. Steady-state methods prioritize computational efficiency and hardware compatibility. Merchant et al.~\cite{merchant2018deep} demonstrated the viability of CNN-based WiFi preamble analysis, attaining 89\% accuracy with 25 Msps sampling rates, thereby balancing precision and resource constraints. Hybrid architectures synergize domain knowledge with adaptive learning. Transformers~\cite{vaswani2017attention} leveraged self-attention for long-range dependency modeling, achieving 91\% accuracy on IEEE 802.11a/g signals. Hybrid frameworks further bridged transient and steady-state paradigms; Han et al.~\cite{Han2025} proposed a VMD-Transformer architecture that decomposes ADS-B signals into multi-frequency IMFs and leverages multi-head self-attention to jointly model high-to-low frequency components. This approach effectively captures both transient spectral distortions and steady-state periodic features, achieving 67.83\% recognition accuracy under mobility-induced Doppler shifts and multipath interference—demonstrating significant robustness in real-world UAV identification tasks.These works collectively validated the integration of signal processing priors with adaptive neural architectures, paving the way for dynamic reconfigurable systems.

\begin{figure*}[t!]
	\centering
	\includegraphics[width=\textwidth]{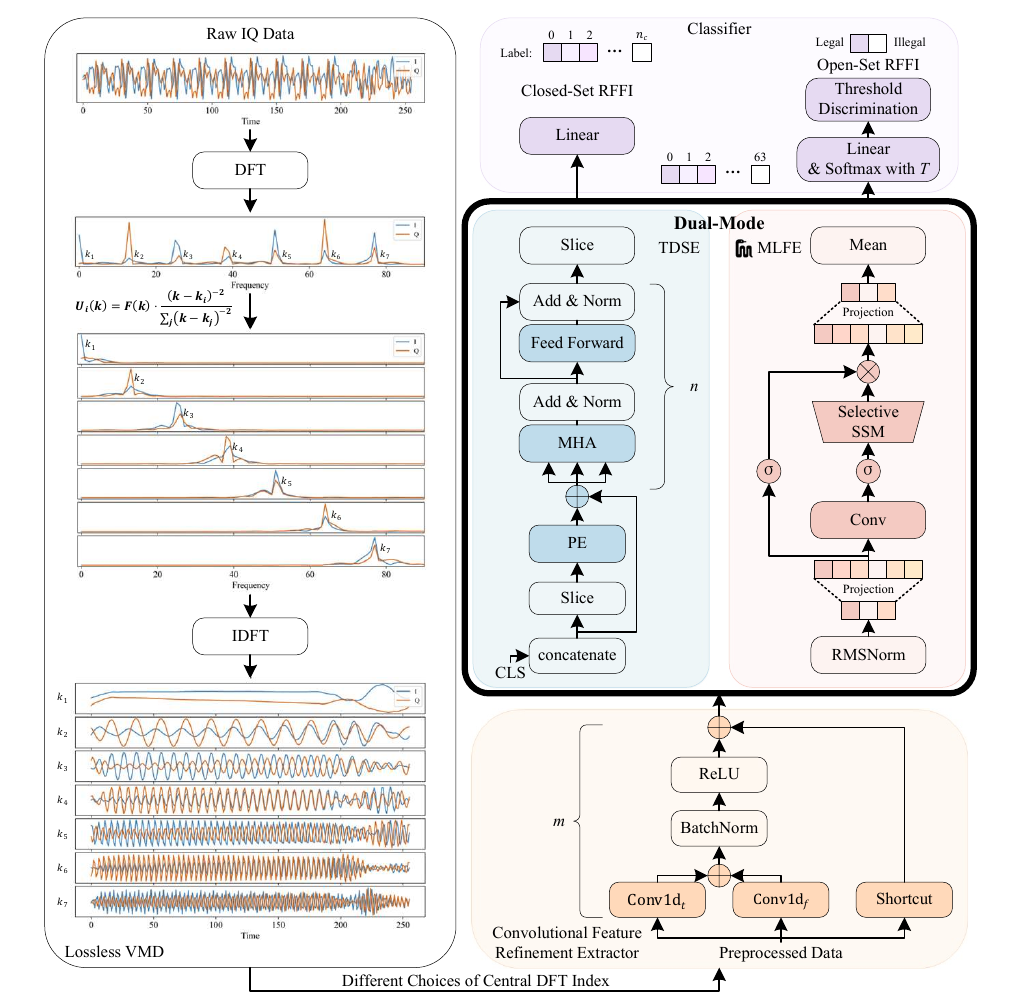}
     \caption{The framework of HyDRA. Input IQ data is processed via lossless VMD, selecting central DFT frequencies as preprocessed data. These are refined by the Convolutional Feature Refinement Extractor (CFRE), encoded by a dual-mode encoder—either the high-precision Transformer Dynamic Sequence Encoder (TDSE) or the efficient Mamba Linear Flow Encoder (MLFE)—and processed through either the Closed-Set RFFI for known RF fingerprint identification or the Open-Set RFFI for detecting unknown transmitters.}
    \label{fig:pipeline}
\end{figure*}

\section{Method}
This section presents HyDRA, a dual-mode framework for radio frequency fingerprinting (RFF) to identify transmitters using IQ data. HyDRA integrates optimized lossless VMD, a dual-mode neural network with Transformer or Mamba encoding, Classifier for either closed- or open-set classification. The framework is illustrated in \cref{fig:pipeline}.

\begin{figure}[htb]
	\centering
	\includegraphics[width=0.4\textwidth]{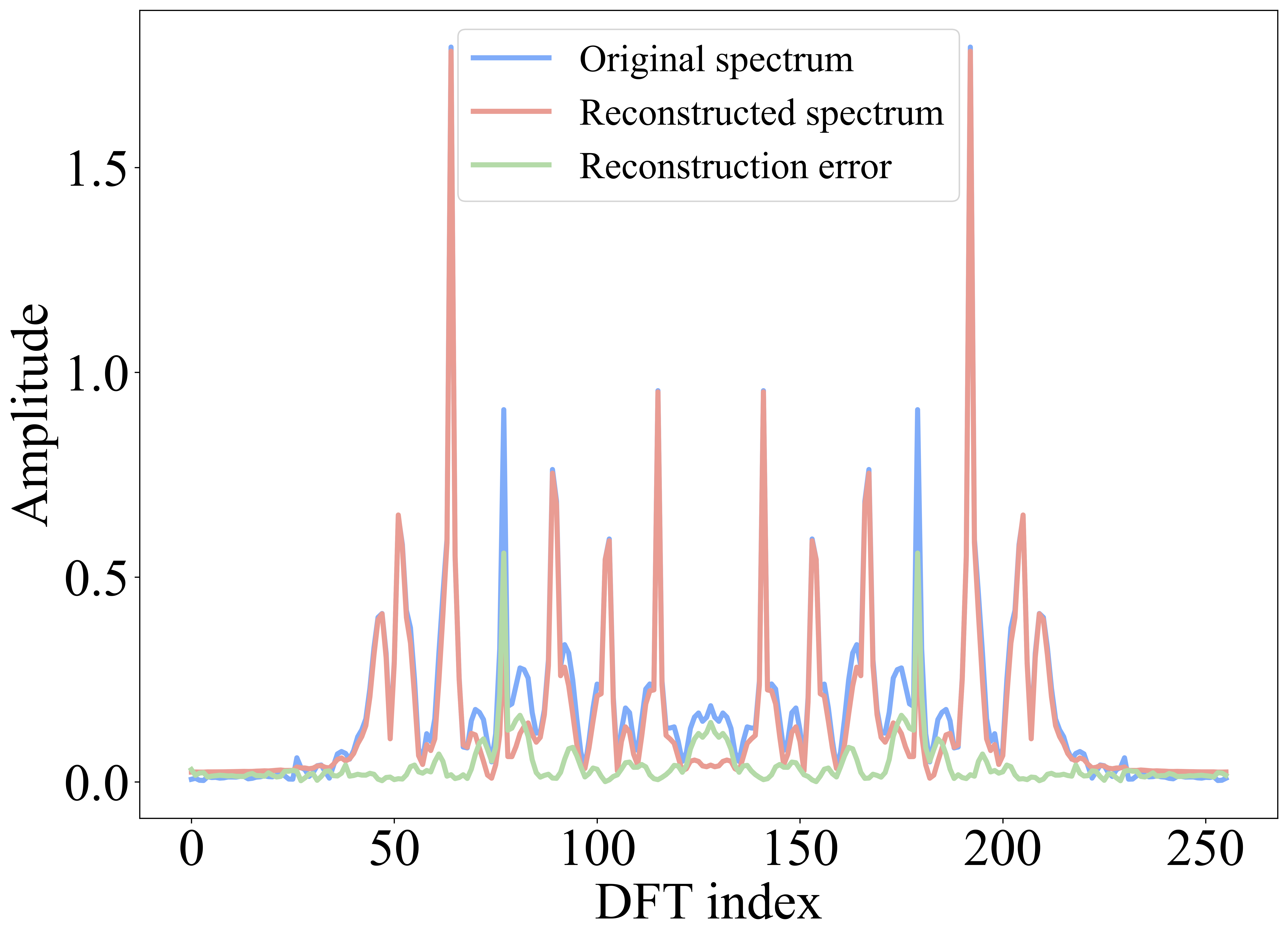}
     \caption{Reconstruction Error of ADMM (k=6).}
    \label{fig:ADMM_error}
\end{figure}

\subsection{Variational Mode Decomposition}
Variational Mode Decomposition (VMD) is an algorithm proposed by Dragomiretskiy et al. \cite{dragomiretskiy2013variational} in 2014, aiming at decomposing a signal into Intrinsic Mode Function (IMF) components,  while ensuring that each component has limited bandwidth. Han et al. \cite{Han2025} have utilized VMD in RFFI and proved that its performance overcomes other mode decomposition methods such as EMD and LMD.

The constrained variational problem of VMD can be expressed as
\begin{equation}
\begin{aligned}
  \argmin{\{u_k\},\{\omega_k\}}\left\{\sum_k \left\|\partial_t\left[\left(\delta(t)+\frac{j}{\pi t}\right)*u_k(t)\right]e^{-j\omega_k t}\right\|_2^2\right\}\\ s.t.\sum_ku_k(t)=f(t) 
\end{aligned}
\label{eq:vmd_t}
\end{equation}
where $f(t)$ is the input signal, $k$ is the number of modes, $u_k(t)$ is the time-domain expression of the $k$-th mode, and $\omega_k(t)$ is the center frequency of the $k$-th mode. $\partial_t$ is the derivative w.r.t. $t$, $*$ denotes the convolution in signal processing and $\|\cdot\|_2$ denotes the $L^2$ norm in functional space $C(-\infty,\infty)$.

By expanding the $L^2$ norm and applying the Parseval's theorem (See Appendix A), the problem can be analyzed in frequency domain, that is
\begin{equation}
\begin{aligned}
  \argmin{\{U_k\},\{\omega_k\}}\left\{\sum_k\int_{0}^{\infty}4(\omega-\omega_k)^2|U_k(\omega)|^2\mathrm{d}\omega\right\} \\ s.t.\sum_kU_k(\omega)=F(\omega) 
\end{aligned}
\label{eq:vmd_f}
\end{equation}
where $U_k(\omega)$ and $F(\omega)$ are the Fourier transform of $u_k(t)$ and $f(t)$, respectively. Evidently, the loss function limits the bandwidth of each mode by penalizing its frequency components far away from its central frequency.

Dragomiretskiy et al. \cite{dragomiretskiy2013variational} introduced an ADMM (Alternate Direction Method of Multipliers) algorithm to solve the optimization problem \cref{eq:vmd_f}. However, this method introduced significant reconstruction error due to relevant terms in the augmented Lagrangian (See \cref{fig:ADMM_error}). This error may serve as denoising at low SNR, but at high SNR, it could be unfavorable for RFFI because it may filter out the crucial spectral features of the transmitter. Furthermore, \cref{fig:ADMM_error} shows that in certain circumstances, the algorithm would locate the central frequencies wrongly As a result, a spectral peak would be strongly suppressed. 

To deal with the drawbacks of ADMM algorithm, we proposed an optimized algorithm based on the closed-form solution of \cref{eq:vmd_f}, shown in \cref{eq:vmd_analytical}. (For mathematical details, see Appendix A).

\begin{equation}
\begin{aligned}
U_k(\omega)=F(\omega)\cdot\frac{(\omega-\omega_k)^{-2}}{\sum_i(\omega-\omega_i)^{-2}}
\end{aligned}
\label{eq:vmd_analytical}
\end{equation}

The center frequencies $\omega_k$ can be determined by solving the optimization problem \cref{eq:vmd_omega}.
\begin{equation}
    \argmin{\{\omega_k\}}\int_{0}^{\infty} \frac{|F(\omega)|^2}{\sum_{k}(\omega-\omega_i)^{-2}}\mathrm{d}\omega
    \label{eq:vmd_omega}
\end{equation}

In practice, the spectrum of a digitally modulated signal reaches a peak at multiples of its fundamental frequency $2\pi/T_s$, where $T_s$ denotes the symbol duration. To suppress the penalty term, the central frequencies of the modes $\{\omega_k\}$ can be directly assigned as multiples of $2\pi/T_s$, without having to solve the optimization problem \cref{eq:vmd_omega}. This approach is more robust, since the central frequencies are manually assigned, eliminating the possibility of location error. In the rest of the paper, this method will be named as \textbf{lossless VMD}, since it does not introduce information loss during decomposition.

To process discrete signals, we transform \cref{eq:vmd_analytical} into its discrete version \cref{eq:vmd_discrete}.
\begin{equation}
\begin{aligned}
U_i(k)=F(k)\cdot\frac{(k-k_i)^{-2}}{\sum_j(k-k_j)^{-2}},\quad k=0,1,\cdots, L-1
\end{aligned}
\label{eq:vmd_discrete}
\end{equation}
where $k_i$ is the corresponding DFT index of the $i$-th central frequency.

With the sampling rate and symbol rate of the dataset known, we can calculate the corresponding DFT index $k_0$ of the fundamental frequency $2\pi/T_s$. As we have discussed, each selected central frequency should be a multiple of $k_0$.
\begin{equation}
    k_0=\frac{f_{sample}}{f_{symbol}}
\end{equation}

VMD can serve as data enhancement by splitting each input into $k$ channels. In our model, the VMD preprocessing layer is a transformation $\phi:\mathbb{R}^{L\times 2}\to \mathbb{R}^{L\times k\times 2}$, mapping the IQ input into VMD-decomposed data.

\subsection{Hybrid Dual-Mode Network Design}
The HyDRA (TDSE/MLFE) is designed for radio frequency fingerprint identification, integrating multiple components to process the input signal and produce the final classification output. Specifically, it comprises the Convolutional Feature Refinement Extractor (CFRE), a dual-mode encoder implemented as either the Transformer Dynamic Sequence Encoder (TDSE) or the Mamba Linear Flow Encoder (MLFE), and the Classifier (including Closed- and Open-Set RFFI). The dual-mode encoder processes \( X_{\text{feat}} \), offering two configurations: a TDSE mode for high precision and an MLFE mode for efficiency. The overall process is mathematically expressed as:
\begin{equation}
X_{\text{feat}} = \text{CFRE}(X),
\end{equation}
\begin{equation}
X_{\text{enc}} = \text{Encoder}(X_{\text{feat}}),
\end{equation}
\begin{equation}
\hat{Y} = \text{Classifier}(X_{\text{enc}}),
\end{equation}
where \( X \in \mathbb{R}^{b \times T \times c} \) is the input tensor, \( X_{\text{feat}} \in \mathbb{R}^{b \times T \times d_m} \) is the feature representation, \( X_{\text{enc}} \) is the encoded sequence (with shape \( \mathbb{R}^{b \times (T+1) \times d_m} \) for TDSE and \( \mathbb{R}^{b \times T \times d_m} \) for MLFE), and \( \hat{Y} \in \mathbb{R}^{b \times n_c} \) is the predicted output, with \( b \) as the batch size, \( T \) as the sequence length, \( c \) as the input channels, \( d_m \) as the feature dimension, and \( n_c \) as the number of classes. Weights for convolutional and linear layers are initialized with \( \mathcal{N}(0, 0.02) \), and batch normalization parameters are initialized with weights of 1 and biases of 0. The following subsubsections detail each component.
\subsubsection{Convolutional Feature Refinement Extractor (CFRE)}
The CFRE module extracts robust features from the input tensor \( X \in \mathbb{R}^{b \times T \times c} \), where \( c \) represents the number of input channels. For IQ data, \( c = 2 \) (in-phase and quadrature components), while for VMD data, \( c = 2k \) (number of variational mode decomposition components). The input is permuted to \( \mathbb{R}^{b \times c \times T} \) for 1D convolution along the temporal dimension. The ResConv1d block, a residual convolutional layer, combines multi-scale temporal features. For an input \( X_{\text{in}} \in \mathbb{R}^{b \times c_{\text{in}} \times T} \), {the block applies a shortcut path, which uses a 1x1 convolution if the input and output channel dimensions differ or an identity mapping otherwise. This is followed by two convolutional operations: one with kernel size \( k_t \) and dilation \( d \) with padding \( p_t = \lfloor k_t / 2 \rfloor \cdot d \), and another with a larger kernel size \( k_f \) and padding \( p_f \). The forward pass proceeds as follows: the shortcut output \( X_{\text{id}} \) is computed from \( X_{\text{in}} \); temporal feature maps \( X_t \) and \( X_f \) are obtained via the respective convolutions; and the block output \( X_{\text{out}} \in \mathbb{R}^{b \times c_{\text{out}} \times T} \) is generated by applying ReLU activation to the batch-normalized sum of \( X_t \) and \( X_f \), then adding \( X_{\text{id}} \). This residual structure aids in capturing both short- and long-range temporal patterns in the RF signals, enhancing the discrimination of subtle device-specific fingerprints by mitigating vanishing gradients and promoting feature reuse.}

{The extractor uses a unified pathway with three ResConv1d layers transforming channels sequentially with increased dilation in the intermediate layer, yielding \( X_{\text{feat}} \in \mathbb{R}^{b \times T \times d_m} \). Overall, the CFRE processes the raw input through these multi-scale convolutions to refine low-level features into higher-dimensional representations that emphasize hardware-induced imperfections, thereby improving the model's ability to distinguish between devices in subsequent encoding stages.}

\subsubsection{Transformer Dynamic Sequence Encoder (TDSE)}
In TDSE mode, implemented as Trans-FluxEncoder, a learnable positional encoding {\( P \in \mathbb{R}^{1 \times T_{\text{max}} \times d_m} \)} is initialized with \( \mathcal{N}(0, 0.02) \), where \( T_{\text{max}} \) is the maximum sequence length. {This encoding is trimmed to match the input sequence length \( T \) and added to the feature representation:}
\begin{equation}
X_{\text{pos}} = X_{\text{feat}} + P,
\end{equation}
enabling the model to learn position-specific patterns. A classification token \( X_c \in \mathbb{R}^{1 \times 1 \times d_m} \), initialized with \( \mathcal{N}(0, 0.02) \), is expanded to \( \mathbb{R}^{b \times 1 \times d_m} \) and concatenated, forming \( X_{\text{seq}} \in \mathbb{R}^{b \times (T+1) \times d_m} \). The sequence is processed by \( L \) layers of a Transformer encoder, each with \( H \) attention heads, feedforward dimension \( d_f \), and dropout \( \delta \), using GELU activation. The multi-head self-attention mechanism, a cornerstone of the Transformer, computes attention for head \( h \) as:
\begin{equation}
\Lambda_h = \text{softmax}\left(\frac{Q_h K_h^\top}{\sqrt{d_k}}\right) V_h,
\end{equation}
where \( Q_h = X_{\text{seq}} W_h^Q \), \( K_h = X_{\text{seq}} W_h^K \), \( V_h = X_{\text{seq}} W_h^V \) are the query, key, and value projections, with \( W_h^Q, W_h^K, W_h^V \in \mathbb{R}^{d_m \times d_k} \), and \( d_k = d_m / H \) is the per-head dimension. The outputs are concatenated and transformed:
\begin{equation}
\text{MultiHead}(X_{\text{seq}}) = \text{Concat}(\Lambda_1, \ldots, \Lambda_H) W^O,
\end{equation}
where \( W^O \in \mathbb{R}^{d_m \times d_m} \) is the output projection. Each layer combines attention with a feedforward network and residual connections, yielding \( X_{\text{enc}} \in \mathbb{R}^{b \times (T+1) \times d_m} \), capturing global dependencies with quadratic complexity in \( T \). {The classification token embedding is then extracted from the first position of the encoded sequence as \( X_{\text{co}} \in \mathbb{R}^{b \times d_m} \) for subsequent prediction.}

{In this encoder, features from the CFRE are further processed through self-attention layers that model long-range interactions across the sequence, allowing the model to integrate contextual information from distant time steps. This enhances discrimination among devices by amplifying unique signal patterns that may be spread out temporally, such as phase distortions or amplitude variations inherent to specific hardware.}

\subsubsection{Mamba Linear Flow Encoder (MLFE)}
In MLFE mode, implemented as Mamba-StreamEncoder, \( X_{\text{feat}} \) is processed by a single Mamba layer, leveraging a selective state-space model (SSM) for linear complexity. The SSM reinterprets sequence modeling as a continuous system, using input-dependent parameters to efficiently capture long-range dependencies. The dynamics for each step \( t = 1, \ldots, T \) are defined as:
\begin{equation}
h_t = \overline{A} h_{t-1} + \overline{B} x_t,
\end{equation}
\begin{equation}
y_t = C h_t,
\end{equation}
where \( x_t \in \mathbb{R}^{b \times d_m} \) is the input at time \( t \), \( h_t \in \mathbb{R}^{b \times d_s} \) is the hidden state, \( y_t \in \mathbb{R}^{b \times d_m} \) is the output, and \( d_s \) is the state dimension. The parameters \( \overline{A} \in \mathbb{R}^{b \times d_s \times d_s} \), \( \overline{B} \in \mathbb{R}^{b \times d_s \times d_m} \), and \( C \in \mathbb{R}^{b \times d_m \times d_s} \) are input-dependent, generated via a linear projection of \( x_t \) to produce \( \Delta, B, C \), where \( \Delta \in \mathbb{R}^{b \times T \times d_m} \) is a timescale parameter (softplus-activated), and discretization yields \( \overline{A} = \exp(\Delta \cdot A) \) and \( \overline{B} = \Delta \cdot B \), with \( A \in \mathbb{R}^{d_s \times d_s} \) as a learnable base matrix. The input is expanded by a factor \( e \), processed with a 1D convolution of kernel size \( k_c \), and selectively gated with SiLU activation, producing \( X_{\text{enc}} \in \mathbb{R}^{b \times T \times d_m} \). This design efficiently captures long-range dependencies, contrasting with the TDSE’s attention-based approach. {The sequence is then mean-pooled over the time dimension to obtain \( X_{\text{co}} \in \mathbb{R}^{b \times d_m} \) for subsequent prediction.}

{Here, the features extracted by the CFRE are streamed through the SSM, which evolves a hidden state over time to accumulate sequential information linearly. This processing refines the features by focusing on temporal evolutions without the computational overhead of attention, thereby improving efficiency while maintaining the ability to discriminate device fingerprints through selective parameter adaptation to input variations.}

\subsubsection{Closed-Set RFFI}
The Closed-Set RFFI generates predictions from \( X_{\text{co}} \). The output is:
\begin{equation}
\hat{Y} = X_{\text{co}} W + b \in \mathbb{R}^{b \times n_c},
\end{equation}
where \( W \in \mathbb{R}^{d_m \times n_c} \) and \( b \in \mathbb{R}^{n_c} \) are learnable parameters, and \( n_c \) is the number of classes.

\begin{figure*}[t!]
	\centering
	\includegraphics[width=\textwidth]{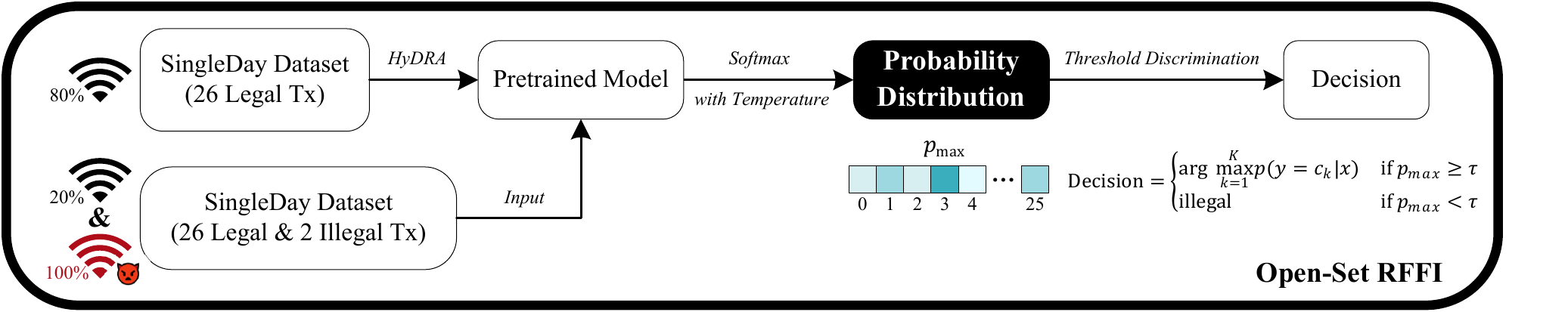}
     \caption{The Open-Set RFFI framework employs the HyDRA (TDSE) model, trained on 80\% of the SingleDay dataset’s 26 legal transmitters (Tx), with the remaining 20\% and 100\% of 2 illegal Tx used for testing. This split balances the proportion of legal and illegal data. The pretrained model outputs a softmax probability distribution over known classes, followed by threshold discrimination to classify device legitimacy.}
    \label{fig_open_set_pipeline}
\end{figure*}

\subsection{Open-Set RFFI}
Current DL-based radio frequency fingerprinting systems predominantly rely on closed-set classification, where models are trained and tested on fixed datasets containing only known devices. In this paradigm, the classifier assumes that all test samples belong to one of the classes observed during training, i.e., \( C = \{c_1, c_2, \ldots, c_K\} \), where \( K \) is the number of known classes. While effective in controlled environments, this approach fails in real-world security applications when unfamiliar or unauthorized devices transmit data. As noted by Al-Hazbi et al.~\cite{open-set}, most existing DL-based RFF systems struggle with new, unseen devices because they have not encountered their unique RF fingerprints during training, leading to unreliable recognition and misclassification based on the closest match from the training set. This limitation underscores the need for open-set classification techniques to enhance system security and adaptability in dynamic, real-world scenarios.

To enhance the security of RFF systems and address the limitations of traditional classification, we developed an algorithm based on open-set classification principles. Unlike closed-set classification, open-set classification accounts for test samples that may belong to unknown classes \( C_{\text{unknown}} \not\subset C \), not seen during training. The algorithm has two primary objectives: (1) accurately classify samples from known classes \( C \) as legal, and (2) identify samples from unknown classes as illegal. This distinction between legal and illegal transmitters strengthens the security of RFF systems for real-world applications. Our approach, depicted in \cref{fig_open_set_pipeline}, leverages the probability distribution from the model's final softmax layer, incorporating a temperature parameter to control the smoothness of the output distribution. For a given input sample \( x \), the softmax layer generates a probability distribution over the known classes, denoted as \( p(y = c_k | x) \) for \( k = 1, 2, \ldots, K \), where \( \sum_{k=1}^K p(y = c_k | x) = 1 \). The softmax function with temperature is defined as:

\begin{equation}
p(y = c_k | x) = \frac{e^{z_k / T}}{\sum_{j=1}^{K} e^{z_j / T}}
\label{eq:softmax with T}
\end{equation}
where \( z_k \) is the raw score (logit) for class \( c_k \), \( T \) is the temperature parameter, \( K \) is the number of known classes, and \( p(y = c_k | x) \) represents the predicted probability of the input \( x \) belonging to class \( c_k \). The temperature parameter \( T \) controls the smoothness of the probability distribution: a higher \( T \) produces a softer, more uniform distribution, while a lower \( T \) sharpens the probabilities, approaching the standard softmax behavior when \( T = 1 \). In a well-trained model, the softmax output for a sample from a known device typically assigns a high probability—often close to 1—to the correct class. For an unknown device, the probabilities are generally lower and more evenly distributed across classes, and the temperature parameter enhances the model's ability to distinguish these cases by adjusting the confidence of the predictions.

We introduce a threshold discrimination mechanism based on the maximum softmax probability, defined as:

\begin{equation}
p_{\text{max}} = \max_{k=1}^K p(y = c_k | x)
\end{equation}

The classification decision is formalized as follows:

\begin{equation}
\text{Decision} =
\begin{cases}
\arg\max_{k=1}^K p(y = c_k | x) & \text{if } p_{\text{max}} \geq \tau \\
\text{illegal} & \text{if } p_{\text{max}} < \tau
\end{cases}
\end{equation}

Here, \( \tau \) is a predefined threshold. If \( p_{\text{max}} \geq \tau \), the sample is classified as belonging to the specific known class given by \( \arg\max_{k=1}^K p(y = c_k | x) \), indicating it likely originates from a recognized device associated with that class. Conversely, if \( p_{\text{max}} < \tau \), the sample is classified as illegal, suggesting it comes from an unknown or unauthorized device. The threshold \( \tau \) serves as a confidence boundary in this threshold discrimination process, separating samples the model recognizes with sufficient certainty for a specific known class from those it does not.

\section{Experiment Setup}
\subsection{Dataset}
{In our experimental setup, we employed two pre-packaged subsets from the WiSig dataset \cite{Wisig2022} – a large-scale, widely adopted\cite{data_ASA}\cite{data_DA}\cite{data_GAN-RXA}\cite{data_OSFS}\cite{data_RFDNA}\cite{data_VC-SEI}\cite{data_ZeroTrust} and recognized collection of WiFi signals specifically designed for receiver- and channel-independent RF fingerprinting.} The SingleDay subset contains 28 transmitters (Tx) and 10 receivers (Rx) captured over a single day, with each Tx-Rx pair providing 800 IEEE 802.11a/g preamble signals at 2 Msps symbol rate (1.0 GB total). The ManyTx subset spans four capture days and includes 150 Tx and 18 Rx, with each Tx-Rx-day combination containing at least 50 signals (2.5 GB total), while tolerating a 10\% imbalance in per-receiver signal counts. Both subsets use equalized preamble signals processed through MMSE channel equalization to mitigate wireless channel effects. We selected these subsets to address distinct experimental requirements: SingleDay facilitates controlled analysis of receiver-specific variations under consistent channel conditions, while ManyTx enables large-scale transmitter discrimination across multiple days and receivers. The dataset's preprocessing ensures compatibility with deep learning workflows by providing standardized 256-sample IQ sequences sampled at 25 Msps. Our choice aligns with the dataset authors' recommendations for benchmarking generalization across hardware and temporal variations. \cref{tab:dataset} concludes the essential information for model training of the two datasets.

\begin{table*}[!ht]
\centering
\small
\setlength{\tabcolsep}{7pt} 
\renewcommand{\arraystretch}{1.2}
\caption{The Accuracies and F1 Scores of Different Methods on SingleDay and ManyTx Datasets. Best Results in \textbf{Bold}.}
\vspace{0.5\baselineskip}
\label{tab:performance_results}
\begin{tabular}{>{\centering\arraybackslash}p{1.5cm}>{\centering\arraybackslash}p{3.5cm}*{4}{c}}
\toprule
\textbf{Dataset} & \textbf{Preprocessing} & \multicolumn{2}{c}{\textbf{HyDRA (TDSE)}} & \multicolumn{2}{c}{\textbf{HyDRA (MLFE)}} \\
\cmidrule(lr){3-4} \cmidrule(lr){5-6}
& & ACC (\%) & F1 (\%) & ACC (\%) & F1 (\%) \\
\midrule
\multirow{3}{*}{\textbf{SingleDay}} & None & $99.77 \pm 0.05$ & $99.77 \pm 0.05$ & $99.68 \pm 0.09$ & $99.68 \pm 0.09$ \\
& VMD (ADMM), k=3 & $99.54 \pm 0.10$ & $99.54 \pm 0.10$ & $99.44 \pm 0.11$ & $99.44 \pm 0.11$ \\
& lossless VMD, k=3 & \bm{$99.96 \pm 0.03$} & \bm{$99.96 \pm 0.03$} & \bm{$99.93 \pm 0.07$} & \bm{$99.93 \pm 0.07$} \\
\midrule
\multirow{3}{*}{\textbf{ManyTx}} & None & $91.81 \pm 0.10$ & $93.04 \pm 0.09$ & $91.50 \pm 0.12$ & $92.64 \pm 0.09$ \\
& VMD (ADMM), k=3 & $89.42 \pm 0.16$ & $90.56 \pm 0.18$ & $87.82 \pm 0.21$ & $89.09 \pm 0.19$ \\
& lossless VMD, k=3 & \bm{$93.97 \pm 0.08$} & \bm{$94.31 \pm 0.07$} & \bm{$92.37 \pm 0.14$} & \bm{$93.25 \pm 0.11$} \\
\bottomrule
\end{tabular}
\end{table*}

\begin{table}[!t]
\caption{Dataset Information}
\vspace{0.5\baselineskip}
\centering
\footnotesize 
\begin{tabular}{@{} l c c c c c @{}} 
\toprule
\textbf{Dataset} & \textbf{Tx No.} & \textbf{Length} & \textbf{Format} & \textbf{Entries} & \textbf{Entries/Tx} \\
\midrule
SingleDay & 28 & 256 & IQ & 224000 & 8000 \\
\addlinespace[0.5ex]
ManyTx & 150 & 256 & IQ & 511515 & 1556--3600 \\
\bottomrule
\end{tabular}
\label{tab:dataset}
\end{table}

\begin{table*}[!t]
\caption{Identification Results of Different Methods on the ManyTx and SingleDay Datasets. Both Datasets have not been Preprocessed. The Best Results Are Highlighted in \textbf{Bold}.}
\vspace{0.5\baselineskip}
\centering
\begin{tabular}{@{} l c c c c @{}}
\toprule
\textbf{Dataset} & \textbf{Method} & \textbf{ACC (\%)} & \textbf{Model Size} & \textbf{Parameter Count} \\
\midrule
\multirow{6}{*}{\textbf{SingleDay}} & CNN\cite{CNN_baseline} & 85 & - & 60,272 \\
& CNN-TFLite\cite{hussain2023edge} & 99 & 462.02 KB  & 116,808 \\
& Transformer-ResNet\cite{transformer_baseline} & 93 & - & - \\
& Transformer-TFLite\cite{hussain2023edge} & 98 & 210.20 KB  & 47,964 \\
& \textbf{HyDRA(TDSE)} & \textbf{99.96} & 600 KB  & 146,876 \\
& \textbf{HyDRA(MLFE)} & \textbf{99.93} & 394 KB & 96,060 \\
\midrule
\multirow{6}{*}{\textbf{ManyTx}} & WiSigNet\cite{Wisig2022} & 53.1 & - & - \\
& WiSigNet MCRFF\cite{MCRFF} & 71.6 & - & - \\
& MobileNet\cite{MobileNetV2} & 84.5 & - & - \\
& MobileNet MCRFF\cite{MCRFF} & 86.4 & - & - \\
& \textbf{HyDRA(TDSE)} & \textbf{93.97} & 600 KB & 146,876 \\
& \textbf{HyDRA(MLFE)} & \textbf{92.37} & 394 KB & 96,060 \\
\bottomrule
\end{tabular}
\vspace*{0.5\baselineskip}
\par
\noindent
\raggedright
{\footnotesize \centering \textit{* All model results used for comparison are sourced from the cited paper.}}
\label{tab:baseline_results}
\end{table*}

\subsection{Baseline Methods}
{To evaluate the advantages of our proposed HyDRA model for RFFI, we selected baseline models evaluated on the same SingleDay and ManyTx datasets, enabling direct and fair performance comparisons. These baselines, encompassing convolutional, contrastive, and transformer-based architectures, employ diverse approaches to RFFI, providing a comprehensive empirical framework to assess HyDRA's performance.}

1) CNN (CNN-based)\cite{CNN_baseline}: This conventional CNN baseline processes raw IQ samples through stacked convolutional layers and pooling operations to extract hierarchical spatial and temporal features from RF signals. It uses convolutional kernels to capture patterns in IQ data and fully connected layers for device classification, focusing on the real and imaginary components to identify unique transmitter characteristics.

2) CNN-TFLite (CNN-based)\cite{hussain2023edge}: This CNN is optimized for edge deployment using TensorFlow Lite (TFLite). It features a streamlined architecture with quantization to reduce model size and latency, targeting real-time RFF on resource-limited devices.

3) Transformer-ResNet (Transformer-based)\cite{transformer_baseline}: This hybrid architecture combines Transformers’ contextual learning with ResNet’s robust feature extraction. It processes IQ samples using the Transformer’s multi-head attention to model long-range dependencies and temporal relationships, while ResNet blocks extract deep, discriminative features, capturing complex, multi-scale RF signal patterns for effective device classification.

4) Transformer-TFLite (Transformer-based)\cite{hussain2023edge}: This method adapts a Transformer encoder for edge use with TFLite optimization. It employs a multi-head self-attention mechanism to analyze IQ samples, focusing on capturing intricate signal relationships in a lightweight format.

5) WiSigNet (CNN-based)\cite{Wisig2022}: WiSigNet, the original convolutional neural network (CNN) proposed by the WiSig dataset authors (denoted as WiSigNet for clarity), is a convolutional neural network (CNN) designed to extract features from raw IQ samples. It relies on a traditional CNN structure to capture spatial patterns in radio frequency signals for device identification.

6) WiSigNet MCRFF (CNN-based)\cite{MCRFF}: An enhanced version of WiSigNet, WiSigNet MCRFF integrates the meta-contrastive learning framework from MCRFF. It enhances feature extraction by optimizing the similarity between representations of seen and unseen devices, leveraging contrastive loss to improve discriminability.

7) MobileNet (CNN-based)\cite{MobileNetV2}: Built on the MobileNetV2 architecture, this lightweight CNN prioritizes computational efficiency. It employs depthwise separable convolutions to process IQ samples, making it suitable for resource-constrained environments while extracting spatial features.

8) MobileNet MCRFF (CNN-based)\cite{MCRFF}: This method integrates MobileNet with the MCRFF framework. By applying meta-contrastive learning, it refines feature extraction to enhance generalization across devices, combining MobileNet’s efficiency with advanced training techniques.

\subsection{Training and Testing Strategies for Closed- and Open-Set RFFI}
This subsection outlines the training and testing strategies for closed-set and open-set Radio Frequency Fingerprint Identification (RFFI) using the HyDRA framework. Each dataset is stratified into three subsets: 80\% for training (\( D_{\text{train}} \)) to optimize model parameters, 10\% for validation (\( D_{\text{val}} \)) to tune hyperparameters and enable early stopping, and 10\% for testing (\( D_{\text{test}} \)) to assess final performance. Stratification preserves class distributions across partitions, maintaining balance in imbalanced scenarios. A deterministic shuffling algorithm with a fixed random seed (42) ensures consistent splits across trials. A ReduceLROnPlateau scheduler adjusts the learning rate when validation loss plateaus, stabilizing training.

For closed-set RFFI, we train and evaluate the model to classify known transmitters using both the SingleDay and ManyTx datasets. These datasets are split as described above, with all samples from known classes used across training, validation, and test sets. This approach assesses the model’s accuracy in identifying registered transmitters within a predefined class set, suitable for controlled environments.

For open-set RFFI, we select the HyDRA (TDSE) model for high accuracy and {divide the SingleDay dataset’s 28 transmitters into 26 legal and 2 illegal ones, consistently selecting the first and second transmitters (IDs `1-11' and `10-11') as the illegal devices} to balance sample sizes. We train and validate the model using only the 26 legal transmitters, following the same 8:1:1 split as closed-set testing, while closed-set testing includes all 28 transmitters. The open-set test set (\( D_{\text{test}} \)) includes test samples from the 26 legal transmitters plus all data from the 2 illegal transmitters, simulating a real-world scenario where unauthorized devices must be detected. This setup, illustrated in \cref{fig_open_set_pipeline}, enables the model to classify legal transmitters into their correct classes while identifying illegal ones as unauthorized. Using the SingleDay dataset ensures data consistency, as all samples are collected on the same day, reflecting realistic operational conditions. Additionally, this approach balances legal and illegal samples with a ratio of approximately 13:10, ensuring a fair evaluation.

In the open-set approach, the model outputs the highest probability across known classes, defined as \( p_{\text{max}} = \max_{k} p(y = c_k | x) \). A threshold \( \tau \) determines classification outcomes:
\begin{itemize}
\item A legal sample is correctly classified if \( p_{\text{max}} \geq \tau \), assigned to the class \( \arg\max_{k=1}^K p(y = c_k | x) \).
\item An illegal sample is correctly classified if \( p_{\text{max}} < \tau \).
\end{itemize}
Formally, with \( N \) total test samples (\( N_{\text{legal}} \) legal and \( N_{\text{illegal}} \) illegal), accuracy is computed as:
\begin{equation}
\text{Accuracy} = \frac{N_{\text{correct, legal}} + N_{\text{correct, illegal}}}{N}
\end{equation}
where \( N_{\text{correct, legal}} \) represents legal samples correctly classified into their classes with \( p_{\text{max}} \geq \tau \), and \( N_{\text{correct, illegal}} \) denotes illegal samples rejected with \( p_{\text{max}} < \tau \). This threshold-based mechanism leverages softmax probabilities to enable robust open-set classification, accurately distinguishing known from unknown transmitters in realistic RFFI applications.

\subsection{Parameter Settings}
The central frequencies of the lossless VMD layer w.r.t $k$ for the proposed HyDRA are summarized in  \cref{tab:ki}. From the spectrum of the signals in the dataset, we observe that the last peak appears at DFT index near $6k_0=75$. For each mode number $k$, we want to choose the central frequencies as uniformly distributed as possible on the $k$-axis, while maintaining each one as a multiple of $k_0$. Here the sample and symbol rate of the dataset are $f_{sample}=25$ Msps and $f_{symbol}=2$ Msps, respectively, resulting in $k_0=12.5$.

\begin{table}[!t]
\caption{Choice of Central DFT Index}
\vspace{0.5\baselineskip}
\centering
\footnotesize 
\begin{tabular}{@{} c c @{}} 
\toprule
\textbf{Mode Number $k$} & \textbf{Central DFT index $k_i$}\\
\midrule
2 & $\{2k_0,4k_0\}$ \\
\addlinespace[0.5ex]
3 & $\{k_0,3k_0,5k_0\}$ \\
\addlinespace[0.5ex]
4 & $\{0,2k_0,4k_0,6k_0\}$ \\
\addlinespace[0.5ex]
5 & $\{0,k_0,3k_0,5k_0,6k_0\}$ \\
\addlinespace[0.5ex]
6 & $\{k_0,2k_0,3k_0,4k_0,5k_0,6k_0\}$ \\
\addlinespace[0.5ex]
7 & $\{0,k_0,2k_0,3k_0,4k_0,5k_0,6k_0\}$ \\
\bottomrule
\end{tabular}
\label{tab:ki}
\end{table}

Training parameters for the proposed HyDRA are summarized in Table~\cref{tab:train_param}. Model training employs the cross-entropy loss function with class weights to address potential imbalances, optimized using the Adam optimizer with an initial learning rate of \( 1 \times 10^{-3} \). A ReduceLROnPlateau scheduler adjusts the learning rate by a factor of 0.1 when validation loss plateaus over 10 epochs, with a minimum learning rate threshold of \( 1 \times 10^{-5} \) triggering early stopping. The batch size is fixed at 32 across all datasets, and training extends up to a maximum of 1000 epochs, with the model achieving the lowest validation loss selected for test evaluation.

\begin{table}[!t]
\caption{Training Parameters}
\vspace{0.5\baselineskip}
\centering
\footnotesize 
\begin{tabular}{@{} l c c @{}} 
\toprule
\textbf{Parameter} & \textbf{Meaning} & \textbf{Value} \\
\midrule
optimizer & & Adam \\
\addlinespace[0.5ex]
$lr$ & initial learning rate & \( 1 \times 10^{-3} \) \\
\addlinespace[0.5ex]
scheduler & & ReduceLROnPlateau \\
\addlinespace[0.5ex]
$\gamma$ & factor of lr reduction & 0.1 \\
\addlinespace[0.5ex]
patience & patience to trigger lr reduction & 10 \\
\addlinespace[0.5ex]
$lr_{min}$ & minimal learning rate & \( 1 \times 10^{-5} \) \\
\addlinespace[0.5ex]
batch size & & 32 \\
\addlinespace[0.5ex]
max epochs & maximum training epochs & 1000 \\
\bottomrule
\end{tabular}
\label{tab:train_param}
\end{table}

Model parameters for the proposed HyDRA, operating in Transformer (TDSE) and Mamba (MLFE) modes, are given in Table~\cref{tab:model_param}. The number of encoding layers varies by mode: \( L_{\text{TDSE}} \) is set to 2 for the TDSE to provide sufficient depth for sequential processing, while \( L_{\text{MLFE}} \) is set to 1 for the MLFE to optimize computational efficiency. The convolutional kernel size in the ResConv1d block is set to \( k_t = 3 \) for one convolutional path and \( k_f = 15 \) for the other, enabling the capture of both localized and broader temporal patterns in the input data. The feature dimension \( d_m \) is configured at 64, balancing representational capacity and computational efficiency. For the TDSE mode, the number of attention heads \( H \) is set to 4, and the feedforward dimension \( d_f \) is 128, optimizing multi-head attention dynamics. In the MLFE mode, the state dimension \( d_s \) of the selective state-space model (SSM) is fixed at 16, the convolutional kernel size \( k_c \) is 4, and the expansion factor \( e \) is 2, enhancing the SSM’s efficiency and expressiveness.

\begin{table}[!t]
\caption{Model Parameters}
\vspace{0.5\baselineskip}
\centering
\footnotesize 
\begin{tabular}{@{} l c c @{}} 
\toprule
\textbf{Parameter} & \textbf{Meaning} & \textbf{Value} \\
\midrule
\( L_{\text{TDSE}} \) & Number of TDSE layers & 2 \\
\addlinespace[0.5ex]
\( L_{\text{MLFE}} \) & Number of MLFE layers & 1 \\
\addlinespace[0.5ex]
\( k_t \) & Temporal convolutional kernel size & 3 \\
\addlinespace[0.5ex]
\( k_f \) & Fixed convolutional kernel size & 15 \\
\addlinespace[0.5ex]
\( d_m \) & Feature dimension & 64 \\
\addlinespace[0.5ex]
\( H \) & Number of attention heads (TDSE) & 4 \\
\addlinespace[0.5ex]
\( d_f \) & Feedforward dimension (TDSE) & 128 \\
\addlinespace[0.5ex]
\( d_s \) & State dimension (MLFE) & 16 \\
\addlinespace[0.5ex]
\( k_c \) & Convolutional kernel size (MLFE) & 4 \\
\addlinespace[0.5ex]
\( e \) & Expansion factor (MLFE) & 2 \\
\bottomrule
\end{tabular}
\label{tab:model_param}
\end{table}

As shown in \cref{tab:Hardware}, the proposed model was trained and tested on an NVIDIA GeForce RTX 3070 Ti with 8 GB of GDDR6 memory, enabling efficient acceleration of the deep learning processes. For real-world deployment, experiments utilized an NVIDIA Jetson Xavier NX, featuring a 6-core Carmel ARM v8.2 CPU and a Volta GPU with 48 Tensor Cores, providing up to 21 TOPS (INT8) of AI performance. Its configurable power consumption (10W/15W/20W) and 8GB LPDDR4x memory with 51.2 GB/s bandwidth support effective real-time inference at the edge with low latency.

\begin{table}[!t]
\caption{Hardware Specifications for Training and Deployment}
\vspace{0.5\baselineskip}
\centering
\footnotesize 
\begin{tabular}{@{} l c c @{}} 
\toprule
\textbf{Property} & \textbf{GeForce RTX 3070Ti} & \textbf{Jetson Xavier NX} \\
\midrule
GPU & NVIDIA GeForce RTX 3070Ti & NVIDIA Volta \\ 
\addlinespace[0.5ex]
CPU & Core i7-8700K & Carmel Arm v8.2 \\ 
\addlinespace[0.5ex]
RAM & 64 GB & 8 GB \\ 
\addlinespace[0.5ex]
Power usage & 240 W & 10 W / 15 W / 20 W \\ 
\addlinespace[0.5ex]
Purpose & Training and testing & Real-world deployment \\ 
\bottomrule
\end{tabular}
\label{tab:Hardware}
\end{table}



\section{Results and Discussions}
\subsection{Performance on Closed-Set RFFI}
Model training is conducted for each architecture with different preprocessing techniques on the SingleDay and ManyTx datasets. {We have performed $k$-fold cross validation with $k=5$ on each combination, and the mean and standard deviation of both accuracy and F1 metrics of the trained models on the test set are listed in Table \ref{tab:performance_results}}. A comparison with baseline methods without preprocessing is provided in \cref{tab:baseline_results}. 
{We observe that all combinations of model architecture and preprocessing techniques exhibit significantly higher performance on the SingleDay dataset than on ManyTx, as the latter has more transmitters, fewer available signal segments for each transmitter (see Table \ref{tab:dataset}), and is collected on 4 different days. On SingleDay, our proposed HyDRA (TDSE), based on a CNN+Transformer architecture, achieves an accuracy of 99.77\% without preprocessing, while HyDRA (MLFE), leveraging a CNN+Mamba architecture, reaches 99.68\%, both delivering state-of-the-art (SOTA) performance that surpasses all baseline methods such as CNN-TFLite (99\%) and CNN (85\%). On ManyTx, HyDRA (TDSE) achieves a remarkable 91.81\% accuracy and HyDRA (MLFE) reaches 91.50\%, likewise attaining SOTA accuracy and outperforming all baselines including MobileNet MCRFF (86.4\%) and WiSigNet (53.1\%), with TDSE slightly ahead of MLFE ($p=0.001$). After adopting our proposed lossless VMD, HyDRA (TDSE) and HyDRA (MLFE) further improve to 99.96\% and 99.93\% on SingleDay, 93.97\% and 92.37\% on ManyTx, respectively, demonstrating their effectiveness across both datasets.}

\subsection{Comparison between ADMM and lossless VMD}
\subsubsection{Accuracy}
To validate the usefulness of the proposed lossless VMD preprocessing method, we controlled the mode number $k=3$ and tested the model accuracy when either VMD (ADMM) or lossless VMD were applied to the TDSE and MLFE architecture. The raw TDSE/MLFE architecture with no preprocessing was set as the control group. {It can be concluded from Table  \ref{tab:performance_results} that for all proposed model architectures, lossless VMD results in higher prediction accuracy compared to the control group, especially in the ManyTx dataset with a 2.16\% improvement. Hypothesis testing has resulted in $p\approx 10^{-10}$, indicating that the performance improvement is statistically significant.} In comparison, VMD (ADMM) performs worse than the control group. This result can prove the claim in Section III. A that the information loss (specifically, loss of crucial spectral features of the transmitter and occasional failure in locating spectral peaks) caused by the nonzero reconstruction error term in the augmented Lagrangian of the ADMM algorithm impairs the model's performance at high SNR scenarios. However, since the lossless VMD method does not introduce information loss, it can improve performance by helping the model recognize different frequency components of the input signal, especially in larger and more complex datasets.

\subsubsection{Execution Time}
\begin{figure}[!t]
\centering
\begin{subfigure}[b]{0.48\columnwidth}
    \includegraphics[width=\textwidth]{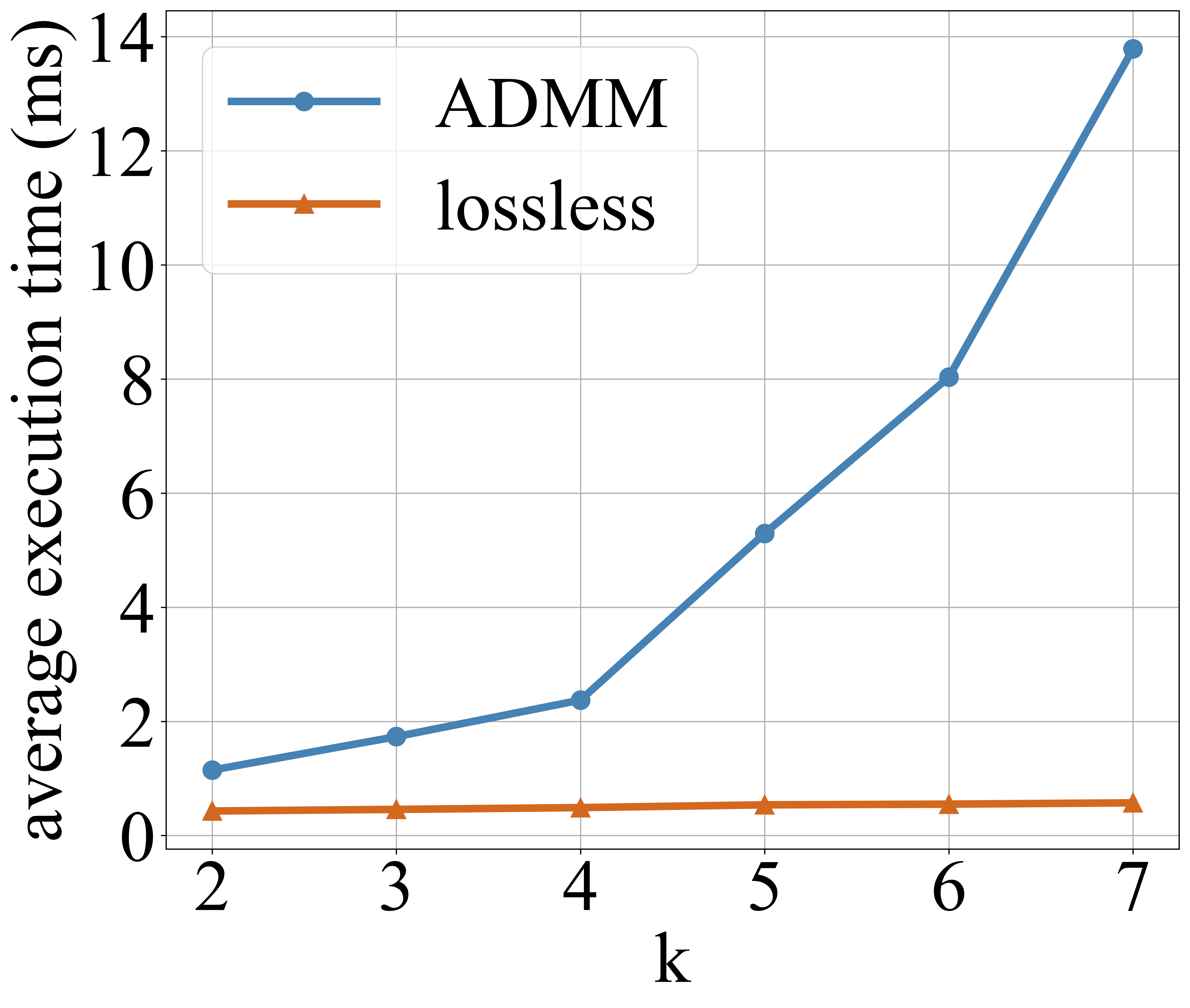}
    \caption{Execution time}
    \label{fig:exec_time}
    \end{subfigure}
\hfill
\begin{subfigure}[b]{0.48\columnwidth}
    \includegraphics[width=\textwidth]{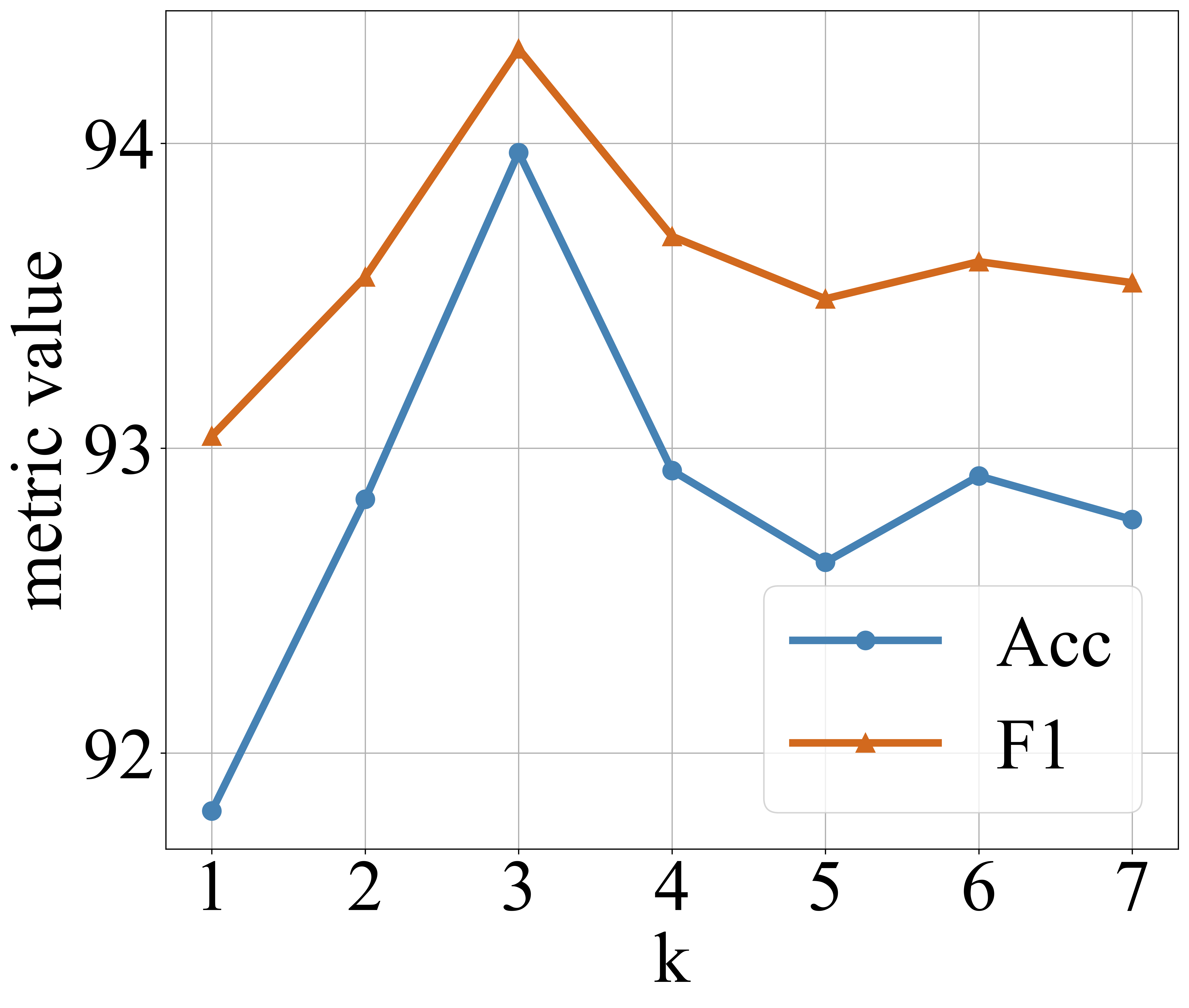}
    \caption{Model accuracy}
    \label{fig:acc_f1_k}
  \end{subfigure}
  \caption{Effect of different $k$ values on the performance of the proposed model. Subfigure (a) compares the average execution time of ADMM and lossless VMD across varying $k$ values. Subfigure (b) illustrates the influence of $k$ on model accuracy and F1-score.}
  \label{fig:loss_acc}
\end{figure}

To analyze the efficiency of the proposed lossless VMD preprocessing method, we compared the execution time of VMD with ADMM (penalty factor set as $\alpha=2000$ proposed in the original paper \cite{dragomiretskiy2013variational}) and lossless VMD. Both methods were performed on 10\% of the entries of the SingleDay dataset (22400 signal segments of length 256) on a Core i7-8700K CPU, and the average execution time was calculated for $k$ between $2$ and $7$, shown in \cref{fig:exec_time}. The figure indicates that VMD (ADMM) executes approximately exponentially long w.r.t. $k$, while the execution time of lossless VMD does not dramatically increase. When $k=3$ (final choice of our model), the average execution time is 0.458ms for lossless VMD and 1.733ms, which is 73.6\% faster than VMD (ADMM). The rate increases w.r.t $k$ and reaches 95.9\% (0.572ms versus 13.785ms) when $k=7$, indicating that lossless VMD has a more significant advantage when decomposing into more IMF components. Since high execution time is unfavorable in real-world applications as it results in high latency, the low time cost is another advantage of lossless VMD.

\begin{figure}[!t]
    \centering
    \includegraphics[width=0.4\textwidth]{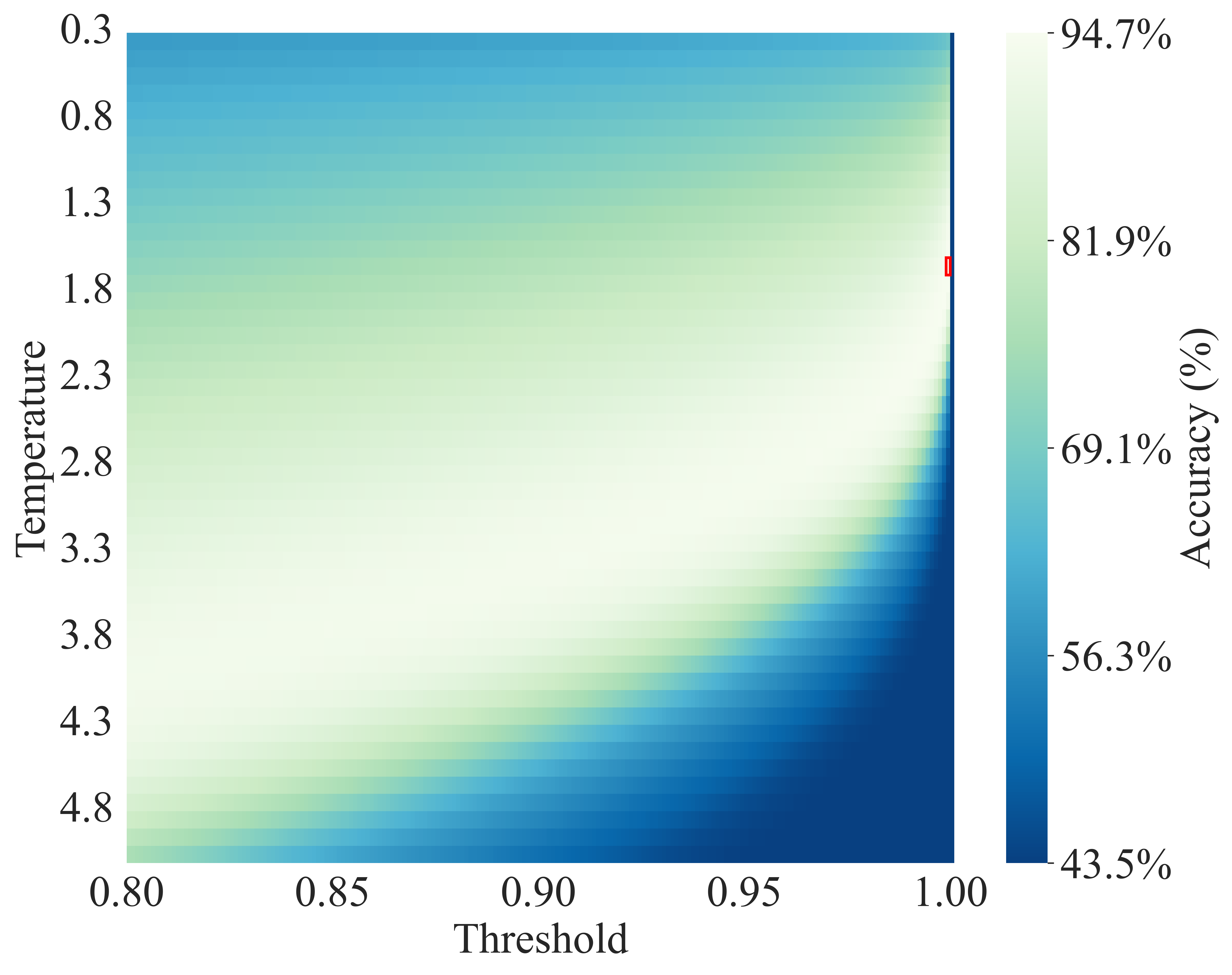}
    \caption{Effect of different threshold and temperature values on Open-Set RFFI accuracy. The highest accuracy is highlighted with a red rectangle.}
    \label{fig:openset}
\end{figure}

\subsection{Effect of the Number of IMF Components}
To analyze the effect of IMF mode number $k$ to the overall performance of the model, we trained the TDSE model on the ManyTx dataset preprocessed by lossless VMD with varying $k$.  {The effect of $k$ on model accuracy and F1 metrics is sketched in Figure \ref{fig:acc_f1_k}, where the range of mode number $k$ from $1$ to $7$ with step size $1$ is selected based on the frequency domain properties of the dataset (7 spectral peaks are detected), covering the scenarios from rough to precise decomposition} ($k=1$ is equivalent to raw IQ data with no preprocessing).

The figure indicates that both accuracy and F1 metric increase and reach a peak of {93.97\%/94.31\%} at $k=3$, and then fall when $k$ continues to increase. This result implies that small or moderate $k$ {(rough or moderate decomposition)} can contribute to the feature extraction of the CFRE layer, while large $k$ {(extremely precise decomposition)} potentially harms the model's ability to generalize. {According to this experiment, we have chosen $k=3$, where the model reaches the peak accuracy, as the final parameter of the lossless VMD layer for our model.}

\subsection{Comparison between HyDRA (TDSE) and HyDRA (MLFE)}
Table~\cref{tab:Comparative} presents a comparison between HyDRA (TDSE) and HyDRA (MLFE), both trained and tested on the SingleDay dataset with lossless VMD preprocessing. HyDRA (TDSE) achieves an accuracy of 99.96\%, slightly ahead of HyDRA (MLFE) at 99.93\% ($p=0.20$), showcasing its strong precision on this dataset. In training time, MLFE stands out with 856.0 seconds, nearly 5.8 times faster than TDSE’s 4929.3 seconds, highlighting its clear efficiency advantage. For model size, MLFE’s 394 KB contrasts with TDSE’s 600 KB, emphasizing MLFE’s suitability for resource-efficient deployment. Both models maintain identical FLOPs of 15,746,816, ensuring consistent computational complexity. The parameter count shows TDSE at 146,876 and MLFE at 96,060, with MLFE’s streamlined design maintaining near-equivalent accuracy at a lower scale, further supporting its efficiency. On the GeForce RTX 3070Ti, MLFE delivers an inference time of 0.3962 ms, slightly faster than TDSE’s 0.4219 ms, reinforcing its speed benefit. These results highlight TDSE’s strength in high accuracy and MLFE’s advantages in training speed, model compactness, and inference efficiency.

\begin{table}[!t]
\caption{Comparison between HyDRA (TDSE) and HyDRA (MLFE)}
\vspace{0.5\baselineskip}
\label{tab:Comparative}
\centering
\footnotesize 
\begin{tabular}{@{} l c c @{}} 
\toprule
\textbf{Property} & \textbf{HyDRA (TDSE)} & \textbf{HyDRA (MLFE)} \\
\midrule
Accuracy & 99.96\% & 99.93\% \\ 
\addlinespace[0.5ex]
Training Time & 4929.3s & 856.0s \\ 
\addlinespace[0.5ex]
Model Size & 600 KB & 394 KB \\ 
\addlinespace[0.5ex]
FLOPs & 15,746,816 & 15,746,816 \\ 
\addlinespace[0.5ex]
Parameter Count & 146,876 & 96,060 \\ 
\addlinespace[0.5ex]
Inf. Time (GeForce RTX 3070Ti) & 0.4219ms & 0.3962ms \\ 
\addlinespace[0.5ex]
Inf. Time (Jetson Xavier NX) & 8.4073ms & 78.2622ms$^*$ \\ 
\bottomrule
\end{tabular}
\vspace*{0.1\baselineskip} 
\par 
\noindent 
\raggedright 
\textit{* Using Mamba-minimal instead of the official Mamba. This does not represent the best outcome.}
\end{table}

\begin{figure}[t!]
    \centering
    \includegraphics[width=0.4\textwidth]{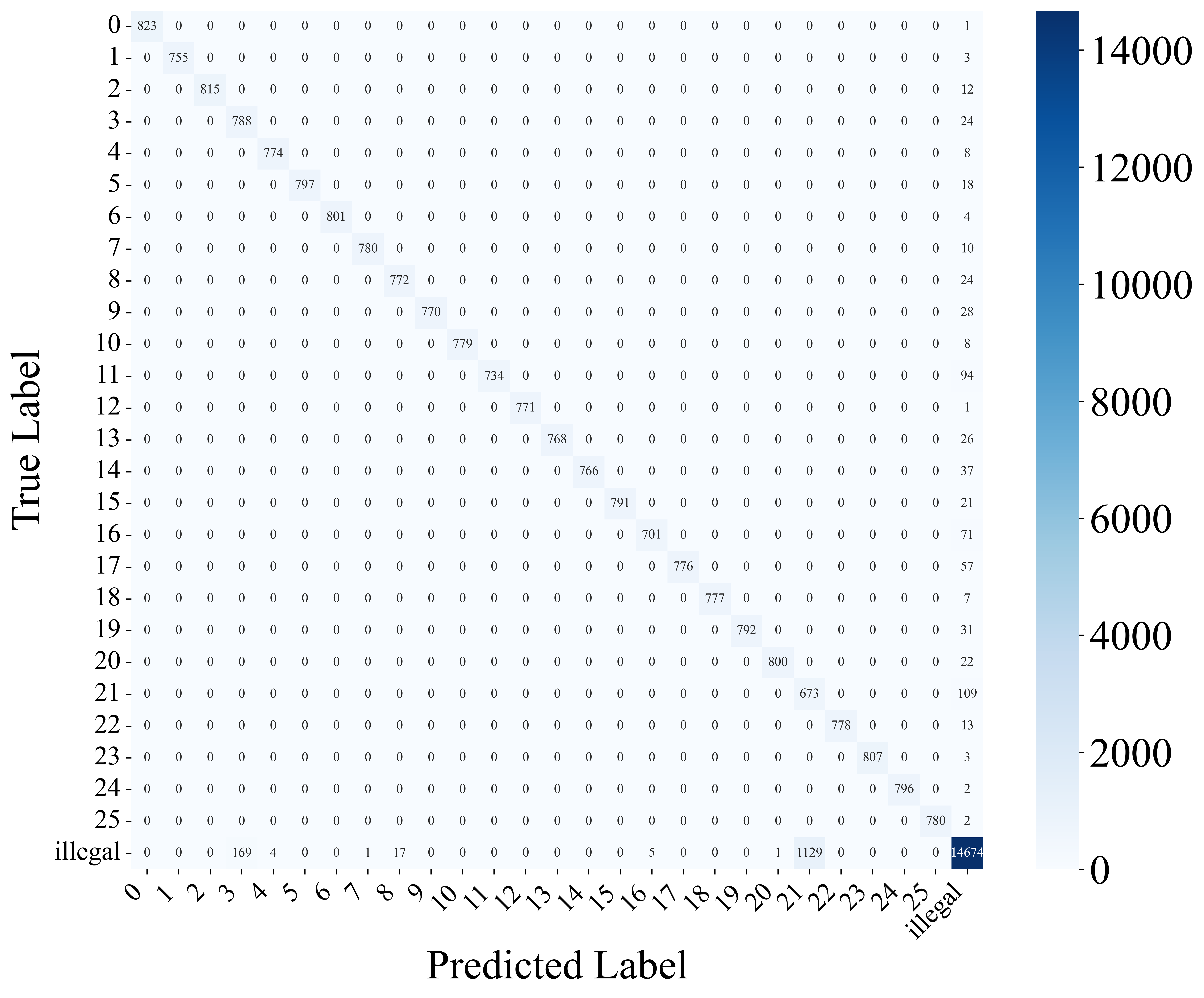}
    \caption{The confusion matrix for Open-Set RFFI at Temperature 1.6 and Threshold 0.999.}
    \label{fig:confusion_matrix}
\end{figure}

\begin{figure}[t!]
    \centering
    \includegraphics[width=0.4\textwidth]{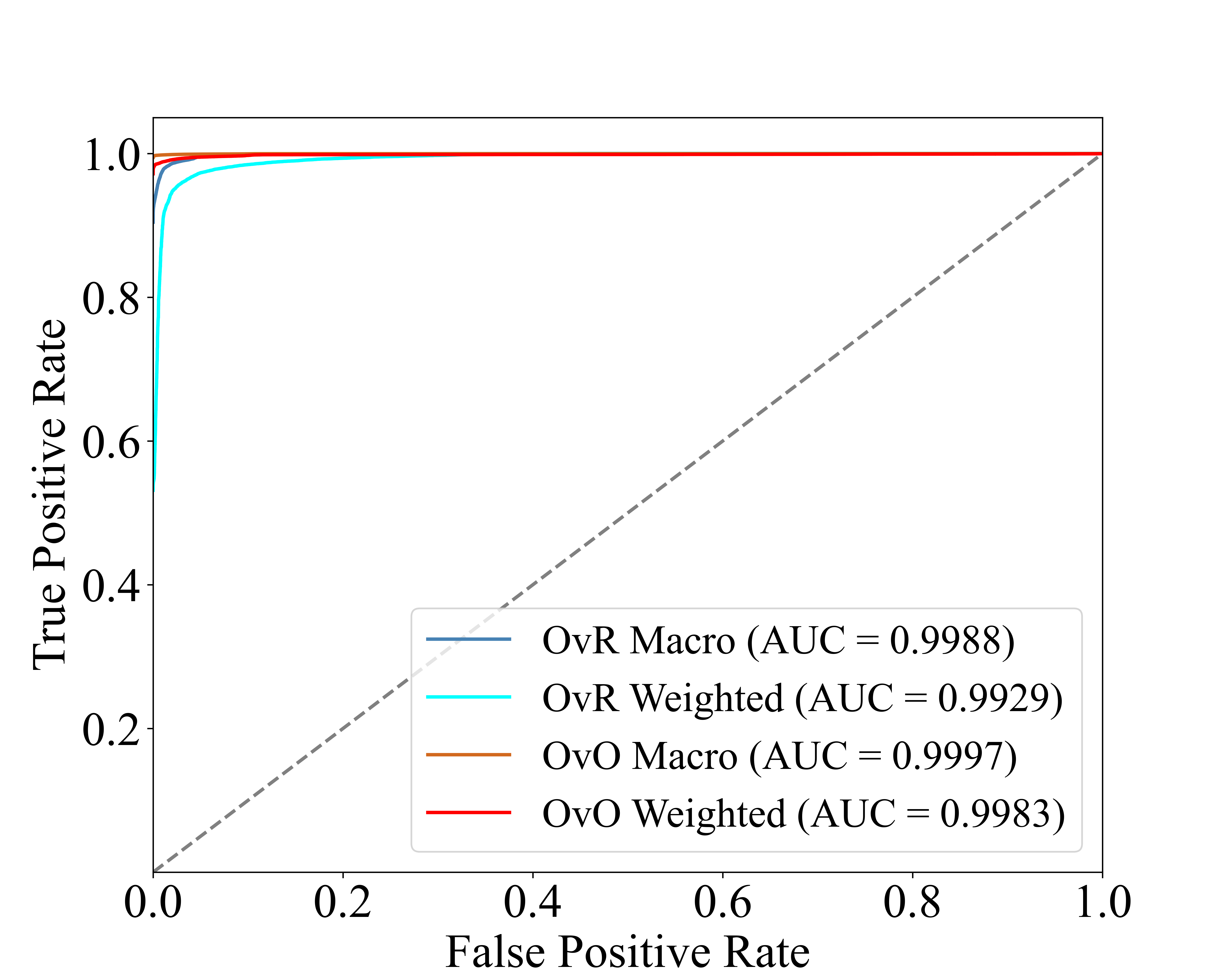}
    \caption{The ROC curves for Open-Set RFFI at Temperature 1.6 and Threshold 0.999}
    \label{fig:roc_auc}
\end{figure}

\begin{table}[!t]
\caption{Evaluation Metrics for Open-Set RFFI at Temperature 1.6 and Threshold 0.999}
\vspace{0.5\baselineskip}
\centering
\begin{tabular}{@{} l c c c c @{}}
\toprule
\textbf{Metric Type} & \textbf{ACC (\%)} & \textbf{Precision (\%)} & \textbf{Recall (\%)} & \textbf{F1 (\%)} \\
\midrule
Macro-Averaged & 94.67 & 96.74 & 96.76 & 96.34 \\
Weighted-Averaged & 94.67 & 96.39 & 94.67 & 95.28 \\
\bottomrule
\end{tabular}
\vspace*{0.5\baselineskip}
\label{tab:metrics_openset}
\end{table}

\subsection{Performance on Open-Set RFFI}
Our algorithm employs the HyDRA (TDSE) model for open-set classification, trained on 26 legal transmitters from the SingleDay dataset, to distinguish legal from illegal devices while detecting unauthorized transmitters in real-world radio frequency fingerprinting scenarios. The effect of varying threshold and temperature parameters on open-set classification accuracy is illustrated in \cref{fig:openset}. {The temperature range of 0.3 to 5.0 with a step size of 0.1 and the threshold range of 0.800 to 1.000 with a step size of 0.001 were selected based on empirical analysis to encompass a spectrum from loose to strict decision boundaries, balancing sensitivity and specificity. These ranges and step sizes were chosen to ensure comprehensive exploration of parameter effects while maintaining computational efficiency. A grid search was conducted to systematically evaluate combinations of these parameters, identifying the optimal pair (temperature = 1.6, threshold = 0.999) that achieves the highest accuracy of 94.67\%, marked by a red rectangle in Figure \ref{fig:openset}. The results demonstrate robustness, with accuracies exceeding 90\% across a wide range of parameter combinations for thresholds below 1, indicating that the model's performance is stable and less sensitive to precise parameter tuning. However, at a threshold of 1, accuracy drops to approximately 45\%, underscoring the importance of parameter optimization.}

Higher thresholds generally require lower temperatures to maintain elevated accuracy levels: as the threshold increases from 0.800 to 0.999, the optimal temperature decreases from approximately 4 to 2. The temperature parameter \( T \) governs the smoothness of the softmax probability distribution, where a higher \( T \) produces a softer, more uniform distribution, and a lower \( T \) sharpens the probabilities. {At higher thresholds, a lower temperature enhances confidence in classifying legal transmitters and rejecting illegal devices}, as described by the temperature-scaled softmax formulation in \cref{eq:softmax with T}. Conversely, setting \( \tau = 1 \) overly sharpens probabilities, leading to {excessive rejections and reduced accuracy.}

{To provide a more comprehensive evaluation, particularly for illegal device detection, we include additional metrics such as false positive rate (FPR), false negative rate (FNR), precision, recall, F1-score, and receiver operating characteristic area under the curve (ROC-AUC), as detailed in Table \ref{tab:metrics_openset} and Figure \ref{fig:roc_auc}. At the optimal parameters, the model demonstrates high accuracy with low false positive rate (FPR) and false negative rate (FNR), effectively minimizing misclassifications of legal devices as illegal and misses of illegal devices. The macro-averaged and weighted-averaged precision, recall, and F1-score, presented in Table \ref{tab:metrics_openset}, indicate balanced performance across classes, with macro-averaging treating all classes equally and weighted-averaging accounting for class imbalance. Additionally, the receiver operating characteristic area under the curve (ROC-AUC) metrics, illustrated in Figure \ref{fig:roc_auc}, showcase the model’s strong discriminative ability in distinguishing legal and illegal transmitters across varying decision thresholds.} The confusion matrix at the optimal temperature of 1.6 and threshold of 0.999, shown in \cref{fig:confusion_matrix}, highlights the model’s performance, correctly identifying most legal transmitters (0 to 25) along the diagonal and classifying a substantial number of illegal devices as illegal, demonstrating robustness in open-set scenarios. However, some illegal devices are misclassified as legal, notably as transmitter 21, suggesting that certain illegal devices may exhibit RF fingerprints similar to legal ones, posing a challenge for differentiation.

{The dataset split, with 26 legal and 2 illegal transmitters (IDs `1-11' and `10-11'), reflects real-world RFFI scenarios where the majority of devices are authorized, and a small number of unauthorized devices must be detected. This configuration ensures that the model is trained on a representative set of legal devices, aligning with practical deployment conditions where known devices dominate. However, the fixed selection of transmitters `1-11' and `10-11' as illegal may limit the model’s exposure to diverse illegal device characteristics, potentially reducing generalizability to unseen unauthorized devices. To further evaluate the impact of dataset splits, we conducted cross-validation experiments, systematically designating each of the 28 transmitters as an illegal device at least once, using the fixed parameters of temperature = 1.6 and threshold = 0.999. This yielded an average accuracy of $93.98\pm0.69\%$, demonstrating the robustness of the method across varied illegal transmitter selections.}

\subsection{Edge Device Deployment}
\begin{figure}[t!]
    \centering
    \includegraphics[width=0.3\textwidth]{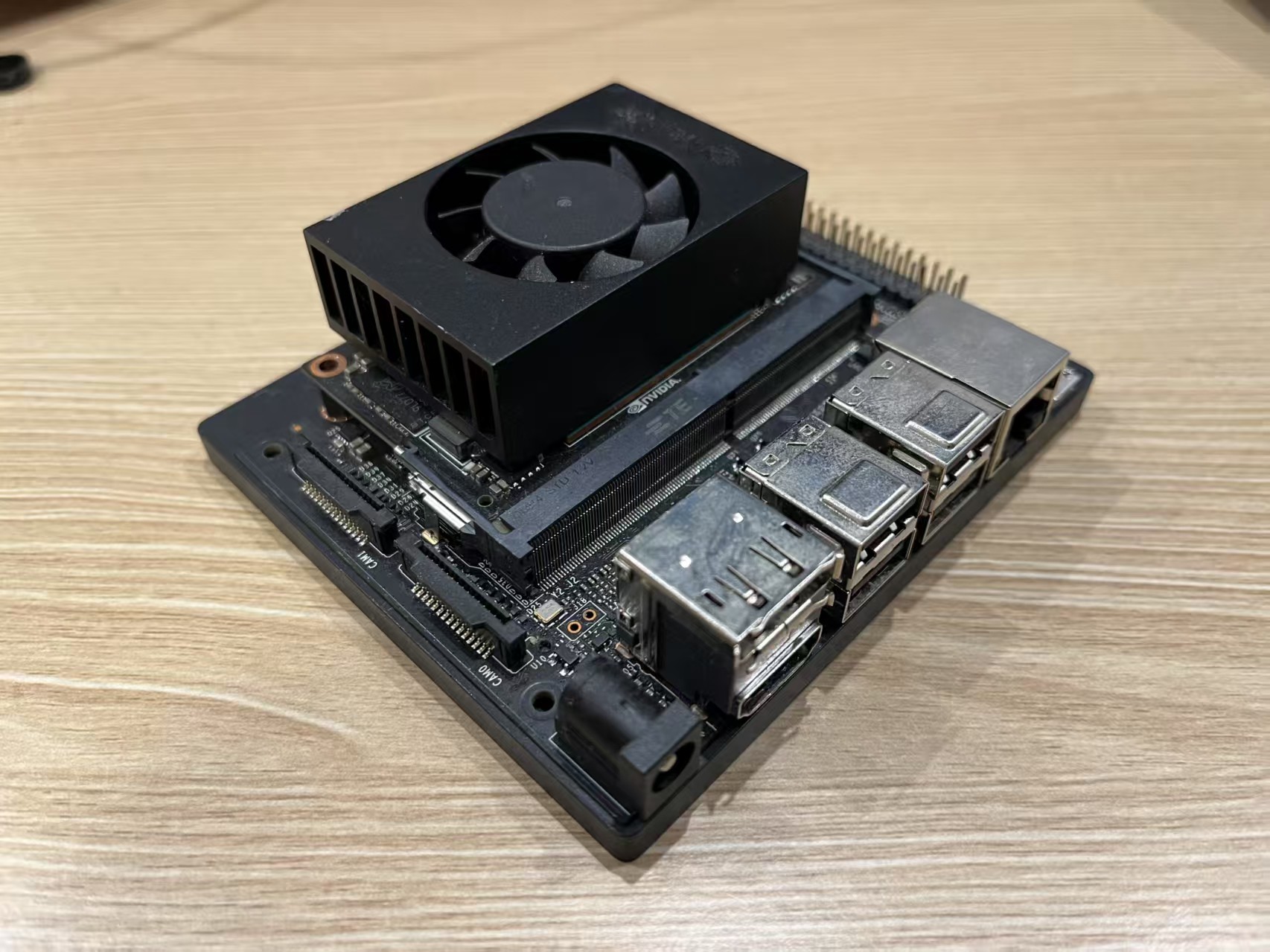}
    \caption{NVIDIA Jetson Xavier NX}
    \label{fig:edge device}
\end{figure}

The HyDRA models, deployed on the NVIDIA Jetson Xavier NX edge computing platform (shown in \cref{fig:edge device}), demonstrate effective performance tailored to their respective configurations, as detailed in Table \cref{tab:Comparative}. To address the incompatibility of the official Mamba library with the edge device’s PyTorch version, we propose a solution by substituting the Mamba core component in HyDRA (MLFE) with Mamba-minimal\footnote{\href{https://github.com/johnma2006/mamba-minimal}{https://github.com/johnma2006/mamba-minimal}}, a streamlined variant of Mamba. This substitution ensures input-output consistency but results in reduced runtime efficiency compared to the official Mamba implementation, which does not represent the optimal outcome. For a detailed assessment of the performance differences between Mamba-minimal and the official Mamba library, refer to the information provided in the Mamba-minimal GitHub repository. HyDRA (TDSE) achieves an inference time of 8.4073 ms, while HyDRA (MLFE) records an inference time of 78.2622 ms. The inference times reported in the table represent the average of 1000 runs following a 10-run warm-up phase. These results underscore the suitability of both models for edge device applications, with HyDRA (TDSE) offering rapid inference and HyDRA (MLFE) benefiting from a reduced parameter count, making them adaptable to diverse data processing needs in resource-constrained settings.

\section{Conclusion}
This paper introduces HyDRA, a Hybrid Dual-mode RF Architecture that enhances Radio Frequency Fingerprint Identification (RFFI) by addressing challenges in dynamic and open-set environments. Unlike traditional methods, HyDRA integrates an optimized Variational Mode Decomposition (VMD) that boosts preprocessing efficiency and accuracy through fixed center frequencies and closed-form solutions. Its adaptable architecture, with a Transformer-based TDSE and a Mamba-based MLFE, delivers robust performance across diverse conditions. Testing on public datasets confirms HyDRA’s exceptional closed-set accuracy and reliable open-set classification, effectively identifying unauthorized devices. Deployment on the NVIDIA Jetson Xavier NX further validates its practicality, achieving millisecond-level inference with low power consumption. {Despite these advances, we acknowledge limitations in real-time signal processing, robustness to environmental variations, and device aging resilience.} {Future work will focus on enabling HyDRA for real-time signal transmission and reception via antenna integration, processing and classifying incoming signals online while determining legitimacy without preprocessed datasets. It will also prioritize enhancing cross-environment generalization through adversarial training with diverse channel conditions, developing aging-aware modeling via incremental learning to track hardware drifts, and integrating real-time anomaly detection to mitigate emerging threats like zero-day spoofing attacks.} These advancements will fortify HyDRA's reliability for mission-critical deployments in volatile real-world settings. This will involve tackling challenges such as real-time signal synchronization, optimizing online VMD processing for low-latency performance, and developing adaptive thresholding for robust open-set classification, ultimately enhancing HyDRA’s applicability in real-time wireless security solutions.

\section*{Acknowledgments}
The authors express their sincere gratitude to Professor Danijela Cabric for her invaluable guidance, insightful suggestions, and continuous support throughout this research. We are particularly thankful for her generous provision of the WiSig dataset, which was instrumental in enhancing the quality and depth of our work.

{\appendices
\section{Mathematical details of lossless vmd}
Let $x_k(t)$ be the analytic, baseband version of $u_k(t)$, i.e.

$$x_k(t)=\partial_t\left[\left(\delta(t)+\frac{j}{\pi t}\right)*u_k(t)\right]e^{-j\omega_k t}$$
\newcommand{\F}[1]{\mathcal{F}\{#1\}}
\newcommand{\sgn}{\mathrm{sgn}}

Perform Fourier transform on both sides.
$$
\begin{aligned}
    X_k(\omega)&=\F{x_k(t)}\\
    &=j\omega\cdot\{[1+\sgn(\omega)]U_k(\omega)*\delta(\omega+\omega_k)\}\\&=j\omega[1+\sgn(\omega+\omega_k)]U_k(\omega+\omega_k)
\end{aligned}
$$

Using the Parseval's relation, the square of its $L^2$ norm is
$$
\begin{aligned}
    \left\|x_k(t)\right\|_2^2&=\int_{-\infty}^{+\infty}|x_k(t)|^2\mathrm{d} t=\int_{-\infty}^{+\infty}|X_k(\omega)|^2\mathrm{d} \omega\\
    &=\int_{-\infty}^{+\infty}|j\omega[1+\sgn(\omega+\omega_k)]U_k(\omega+\omega_k)|^2\mathrm{d} \omega\\
    &=\int_{-\infty}^{+\infty}|j(\omega-\omega_k)[1+\sgn(\omega)]U_k(\omega)|^2\mathrm{d} \omega\\
    &=\int_{0}^{+\infty}4(\omega-\omega_k)^2|U_k(\omega)|^2\mathrm{d} \omega
\end{aligned}
$$

Plug it into \cref{eq:vmd_t} we derive \cref{eq:vmd_f}. 

When $\omega_k$ is fixed, we consider each $\omega\geq0$ separately, since the integrated term does not depend on other frequency components. The optimization problem thus turns into

$$
\begin{aligned}
  \argmin{\{U_k(\omega)\}}\left\{\sum_k4(\omega-\omega_k)^2|U_k(\omega)|^2\right\} \\ s.t.\sum_kU_k(\omega)=F(\omega) 
\end{aligned}
\label{eq:vmd_f_separate}
$$

Let $\beta_k=4(\omega-\omega_k)^2\geq 0$, $\alpha_k=U_k(\omega)/F(\omega)$, we rewrite the problem as
$$
\argmin{\{\alpha_k\}}\left\{\sum_k\beta_k|\alpha_k|^2\right\} \quad s.t.\sum_k\alpha_k=1 
$$

It is evident that $\alpha_k$ must be real, or we can replace $\alpha_k$ with $Re(\alpha_k)$ to decrease $L(\boldsymbol{\alpha})=\sum_k\beta_k|\alpha_k|^2$. 

Using the Cauchy-Schwartz inequality, we have
$$
\sum_k\beta_k\alpha_k^2\cdot\sum_k\beta_k^{-1}\geq\left(\sum_k\alpha_k\right)^2=1
$$
and the equality holds when $\alpha_k/\beta_k$ is a constant, i.e, $\alpha_k\subset (\omega-\omega_k)^2$, which derives \cref{eq:vmd_analytical}. Plugging \cref{eq:vmd_analytical} into \cref{eq:vmd_f} gives \cref{eq:vmd_omega}. 
}

\footnotesize
\normalem
\bibliography{ref}
\bibliographystyle{IEEEtran}

\end{document}